\documentclass[fullpaper,final]{nldl}

\paperID{42}
\confYear{2024}

\vol{233}

\usepackage{mathtools}
\usepackage{amsfonts}

\usepackage{graphicx}

\usepackage{booktabs}

\usepackage{enumitem}

\usepackage{algorithm}
\usepackage{algorithmic}

\usepackage{listings}
\usepackage{hyperref}
\usepackage{url}
\hypersetup{
  pdfusetitle,
  colorlinks,
  linkcolor = BrickRed,
  citecolor = NavyBlue,
  urlcolor  = Magenta!80!black,
}

\title{Guiding drones by information gain}
\author[1]{Alouette van Hove\thanks{Corresponding Author.}}
\author[1]{Kristoffer Aalstad}
\author[1]{Norbert Pirk}
\affil[1]{Department of Geosciences, University of Oslo, Norway}
\affil[ ]{\texttt{a.van.hove@geo.uio.no}}

\begin{document}
\maketitle

\begin{abstract}
The accurate estimation of locations and emission rates of gas sources is crucial across various domains, including environmental monitoring and greenhouse gas emission analysis. This study investigates two drone sampling strategies for inferring source term parameters of gas plumes from atmospheric measurements. Both strategies are guided by the goal of maximizing information gain attained from observations at sequential locations. Our research compares the myopic approach of infotaxis to a far-sighted navigation strategy trained through deep reinforcement learning. We demonstrate the superior performance of deep reinforcement learning over infotaxis in environments with non-isotropic gas plumes. 
\end{abstract}

\section{Introduction}

As a result of global warming, permafrost soils are thawing and will potentially release large amounts of greenhouse gases such as methane and carbon dioxide into the atmosphere~\cite{schuur_climate_2015}. Understanding the scale and distribution of these greenhouse gas sources helps improve the reliability of climate projections. 

Determining the location and flux (i.e. emission rate) of greenhouse gas sources remains challenging as they cannot be observed directly. Instead, these source term parameters must be inferred through the analysis of observational data, such as gas concentrations and other atmospheric variables. This inverse problem is known as source term estimation~(STE)~\cite{hutchinson_review_2017}. The Bayesian inversion approach combines observational data with an atmospheric plume model, accounting for uncertainties in both, to yield a probability distribution over possible source term parameters. Sequentially applying Bayes' theorem to each observation updates the parameters' probability distribution~\cite{Sarkka_2023} so as to constrain uncertainty in the source term estimates. 

Observations can be acquired using static sensors (e.g. measurement towers), mobile sensors (e.g. drones), or a combination of the two. Although mobile sensors offer the advantage of flexibility and adaptability, the challenges of devising the optimal sampling strategy become even more apparent. Sample trajectories can be pre-planned in an offline manner, for example, by selecting a sweeping ``lawnmower'' pattern, or they can be adapted to the environment in an online manner~\cite{francis_gas_2022}. The latter is known as informative path planning~(IPP). 

In this paper, we explore two approaches to IPP for addressing the STE problem. The first approach involves a modified version of infotaxis, an online search strategy initially introduced by \citeauthor{vergassola_infotaxis_2007}~\cite{vergassola_infotaxis_2007} for source localisation. Our work extends this algorithm to estimate multiple source term parameters. The second approach uses deep reinforcement learning~(DRL)~\cite{sutton_reinforcement_2018} to develop a sampling strategy. We train a mobile drone-based sensor to obtain the most certain estimates of the source term parameters within a finite number of observations, simulating the constraints of a drone's limited battery life. This study aims to determine whether a DRL strategy with an information-based reward can outperform the infotaxis strategy in the context of STE.

Recent studies have explored the application of DRL for guiding mobile sensors in source seeking tasks~\cite{duisterhof_learning_2021, loisy_searching_2022, loisy_deep_2023, wang_olfactory-based_2021, zhao_deep_2022}. However, these studies focused solely on source localisation, neglecting source strength estimation. Moreover, task completion in \cite{duisterhof_learning_2021, loisy_searching_2022, loisy_deep_2023, zhao_deep_2022} was determined based on the proximity of the mobile sensor to the source. This is impractical for the real-world identification of unknown greenhouse gas sources whose location is uncertain. This study seeks to jointly address both the source localisation and strength problem. While~\citeauthor{park_source_2022}~\cite{park_source_2022} did consider the STE problem, they aimed to minimize the number of observations needed for accurate STE, rather than minimizing uncertainty within a single sampling flight. In line with \citeauthor{vanHove2023}~\cite{vanHove2023}, we apply an information-based reward function. Whereas this earlier study estimated the source strength at a known location using tabular RL, we apply deep RL herein to account for the larger parameter space. 

The study's primary contributions are as follows: (a)~Application of DRL to address the STE problem under pertinent constraints for greenhouse gas mapping in the field - considering unknown source locations, fixed flight time due to battery limitations, and noisy observations by the sensor. (b)~Comparing the performance of DRL with an information-based reward to infotaxis. (c)~Comparison between two neural network architectures: a feed forward structure with fully connected layers (as utilized by~\cite{duisterhof_learning_2021, loisy_searching_2022, loisy_deep_2023, wang_olfactory-based_2021, zhao_deep_2022}) and a convolutional neural network. 

\section{Problem description}~\label{sec:problem_description} 

The navigation of mobile sensors for STE is a sequential decision-making problem characterized by uncertainty and partial observability. The mobile sensor cannot directly perceive the true state of the emissions but instead receives incomplete and uncertain observations related to gas concentration. In a mathematical framework, the problem can be represented as a Partially Observable Markov Decision Process~(POMDP)~\cite{kaelbling_planning_1998}, that can be framed as a belief Markov Decision Process (belief-MDP) where states are replaced by belief states. In this context, the mobile sensor - \emph{the agent} -  maintains a probability distribution over the unknown source term parameters - \emph{its belief} - while navigating a two-dimensional grid. At each time step, the agent can choose from a set of five actions: \{$\rightarrow$, $\leftarrow$, $\downarrow$, $\uparrow$, stay\}. After executing an action, the agent receives a noisy observation which it uses to update its beliefs about the source term parameters through Bayesian inference. Solving the navigation problem for STE with mobile sensors means finding the optimal strategy - \emph{the policy} - for mapping belief states to actions. The agent's objective is to maximize a reward function that incentivizes it to select actions leading to more certain source term estimations.

\subsection{Observation model}~\label{sec:observation_model}

\begin{figure}[tb]
    \centering
    \includegraphics[width=0.9\linewidth]{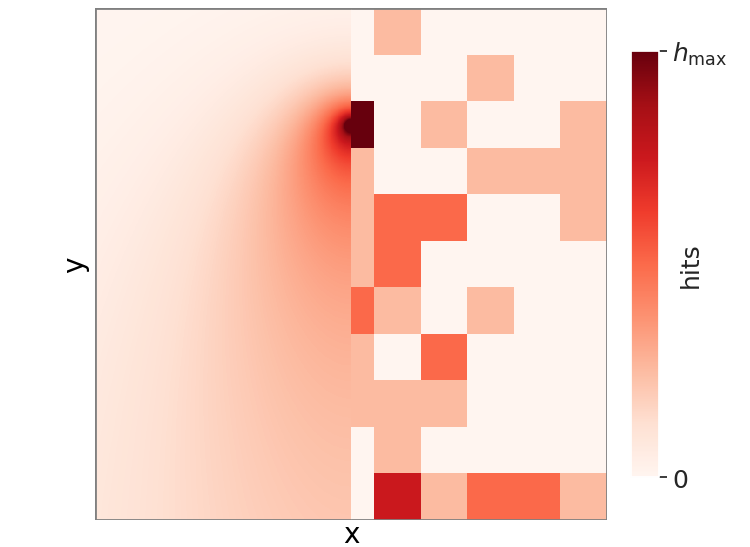}
    \caption{Example of a hit map: (left half) map of the mean number of hits~$\mu$ calculated by the analytic plume model Eq.~\eqref{eq:plume_model}, and (right half) a possible map of noisy sensor measurements~$h$ from Eq.~\eqref{eq:poisson}.}
    \label{fig:hit_field}
\end{figure}

A greenhouse gas source emits detectable particles that are dispersed through domain $\mathcal{A}$ under turbulent transport conditions. We use the advection-diffusion plume model presented in~\cite{vergassola_infotaxis_2007} to compute the mean number of detected particles, or ``hits'', by an agent at position $\mathbf{x} = \left[ x, y \right] \in \mathcal{A}$ during a sampling time interval~$\Delta t$. The source term parameter vector $\boldsymbol{\theta}$ includes the source location $\mathbf{x}_{\mathrm{s}} = \left[ x_{\mathrm{s}}, y_{\mathrm{s}} \right] \in \mathcal{A}$ and its source strength (i.e. flux) $\phi \in \mathbb{R}^{+}$. In two dimensions, the mean number of hits detected by an agent with body radius~$r$ is given by Eq.~\eqref{eq:plume_model}
\begin{equation} \label{eq:plume_model}
\begin{split} 
    \mu( \boldsymbol{\theta}, \mathbf{x}) &= \phi \Delta t \frac{1}{\ln( \frac{\lambda}{r})} \exp \left( \frac{-(y - y_{\mathrm{s}})V}{2D} \right) \\
    & K_0 \left( \frac{ \lVert \mathbf{x}-\mathbf{x}_{\mathrm{s}}\rVert }{\lambda} \right) \, ,
\end{split}
\end{equation}
where the wind has been taken to blow in the negative $y$-direction. $K_0$ is the modified Bessel function of the second kind of order 0. In this model, the gas particles have a finite lifetime~$\tau$. Mean wind speed~$V$ and effective diffusivity~$D$ (i.e. turbulent and molecular diffusivity) affect the dispersion length scale~$\lambda$ of the particles by Eq.~\eqref{eq:lambda}
\begin{equation} \label{eq:lambda}
    \lambda = \sqrt{ \frac{D \tau}{1 + \frac{V^2 \tau}{4D} } }\, .
\end{equation}
An example of a mean hit map is shown on the left half of Figure~\ref{fig:hit_field}. The maximum number of hits is fixed to an upper bound, $h_{\textrm{max}}$. The mean hit map becomes isotropic in the absence of wind.

The stochastic data generating process for the number of hits detected by the sensor during sampling interval~$\Delta t$ is modelled by a Poisson distribution. The probability that the sensor detects $h$ hits is given by Eq.~\eqref{eq:poisson}
\begin{equation} \label{eq:poisson}
    \Pr (h | \mu ) = \frac{ \mu^h }{ h! } \exp(-\mu) \, .
\end{equation}

The right half of Figure~\ref{fig:hit_field} shows a possible measurement map. The difference between the left half and right half of the map demonstrates the challenge of STE in turbulent conditions. Observations are sparse and noisy, and gradient-based sampling strategies for STE are therefore often not optimal and can fail in turbulent flows. 

\subsection{Bayesian inference}~\label{sec:bayesian_inference}

Each hit observation provides information about the source term parameters. The agent applies Bayesian inference after each observation to maintain an up-to-date belief about the source term parameters $\Pr(\boldsymbol{\theta})$, where $\boldsymbol{\theta}$ is a discrete random variable of all possible source locations and fluxes. In particular, Bayes' rule~\eqref{eq:Bayes} maps the prior belief $\Pr(\boldsymbol{\theta}_{n})$ to the posterior belief $\Pr(\boldsymbol{\theta}_{n}|h_{n})$ after a hit observation $h_{n}$ at time $t_{n}$ 
\begin{equation}
\Pr(\boldsymbol{\theta}_{n}|h_{n})=\frac{\Pr(h_{n}|\boldsymbol{\theta}_{n})\Pr(\boldsymbol{\theta}_{n})}{\Pr(h_{n})} \, , \label{eq:Bayes}
\end{equation}
where the likelihood $\Pr(h_n|\boldsymbol{\theta}_n)$ is computed by Eq.~\eqref{eq:poisson} with $\mu$ given by Eq.~\eqref{eq:plume_model}. All probabilities are implicitly conditioned on the plume model and background information, including agent's location~$\mathbf{x}$ and wind speed~$V$, among others. The evidence $\Pr(h_n)=\sum_i \Pr(h_n|\boldsymbol{\theta}_n^{(i)})\Pr(\boldsymbol{\theta}_n^{(i)})$ acts as a normalizing constant such that the posterior belief over all possible $\boldsymbol{\theta}_n^{(i)}$ sums to one. Our agent applies Eq.~\eqref{eq:Bayes} sequentially in time by letting the posterior at time $t_n$ become the prior at time $t_{n+1}$, $\Pr(\boldsymbol{\theta}_{n+1})=\Pr(\boldsymbol{\theta}_n|h_n)$, resulting in dynamic belief updating. 

\section{Sampling strategies}~\label{sec:sampling_strategies} 

Within the belief-MDP framework, the agent's belief state is denoted as $s = (\mathbf{x}, \Pr(\boldsymbol{\theta}))$. In this study, both DRL and infotaxis policies utilize a reward function based on information entropy. Information entropy~$H$~\cite{mackay_information_2003} is non-negative and measures the uncertainty of a probability distribution. The information entropy (measured in $\textrm{nats}$) of a belief state~$s$ is given by Eq.~\eqref{eq:entropy}
\begin{equation} \label{eq:entropy}
    H(s) = - \sum_i \Pr(\boldsymbol{\theta}^{(i)}) \ln \Pr(\boldsymbol{\theta}^{(i)}) \, .
\end{equation}
Note that information entropy of a belief state is not directly affected by the agent's location~$\mathbf{x}$, and relies only on $\Pr(\boldsymbol{\theta})$, which in turn depends on the agent's sampling strategy and observations. The initial prior belief of the agent is a uniform distribution over all possible source locations and fluxes, resulting in maximum information entropy. Conversely, the information entropy of a Kronecker delta distribution, where the probability is one for a specific $\boldsymbol{\theta}^{(i)}$ and zero for all others, attains the minimum information entropy of zero. 

\subsection{Infotaxis}~\label{sec:infotaxis}

Infotaxis~\cite{vergassola_infotaxis_2007} is a greedy policy which sequentially maximizes its reward function. At each step, the agent selects an action~$a$ that maximizes the expected information gain~$G$~\cite{loisy_searching_2022}, such that Eq.~\eqref{eq:infotaxis_policy} follows
\begin{equation} \label{eq:infotaxis_policy}
    \pi_{\textrm{infotaxis}} (s) = \arg\max_a G(s, a) \, .
\end{equation}
Expected information gain is the expected difference in information entropy between the agent's current belief state $s$ and all possible successor belief states $s'$, given by Eq.~\eqref{eq:information_gain}
\begin{equation} \label{eq:information_gain}
    G(s, a) = H(s) - \sum_{s'} \Pr(s'| s, a) H(s') \, .
\end{equation}
The agent can reach different successor belief states depending on the hits it receives after performing action~$a$. Combining Eq.~\eqref{eq:infotaxis_policy} and \eqref{eq:information_gain} gives Eq.~\eqref{eq:infotaxis_policy_2}
\begin{equation} \label{eq:infotaxis_policy_2}
    \pi_{\textrm{infotaxis}} (s) = \arg\min_a \sum_{s'} \Pr(s'| s, a) H(s') \, .
\end{equation}
such that the policy can also be formulated as selecting the action that transitions the agent to the successor belief state with the minimum expected information entropy (i.e. minimum uncertainty). 

\subsection{Deep reinforcement learning}~\label{sec:DRL}

Infotaxis is a myopic approach, aiming to maximize the expected reward of the next action. In contrast, RL trains a policy to maximize the cumulative reward obtained throughout an entire \emph{episode}, considering a sequence of actions. The essence of RL lies in identifying the optimal policy associated with the optimal value function~$v_{*}(s)$. This represents the maximum expected \emph{cumulative} reward that can be achieved from a given belief state. The large number of possible beliefs $\Pr(\boldsymbol{\theta})$ in a belief state~$s$ prevents the use of exact methods for obtaining $v_{*}(s)$. Consequently, we approximate $v_{*}(s)$ through a parameterized function $\hat{v}(s, \mathbf{w})$, in the form of a deep neural network~\cite{kochenderfer_algorithms_2022} with weight vector~$\mathbf{w}$. The Bellman optimality equation for the approximate value function~\cite{sutton_reinforcement_2018} is given by Eq.~\eqref{eq:general_bellman_optimality} 
\begin{equation} \label{eq:general_bellman_optimality}
    \hat{v}(s, \mathbf{w}_{*}) = \max_a \sum_{s',r} \Pr(s',r|s,a) \left[ r + \hat{v}(s', \mathbf{w}_{*}) \right] \, ,
\end{equation}
where $r$ is the reward. We use a reward function based on information gain~\eqref{eq:information_gain}: $r = - H(s')$. This reward function depends only on successor state $s'$, such that Eq.~\eqref{eq:bellman_optimality} follows as the Bellman optimality equation for this study
\begin{equation} \label{eq:bellman_optimality}
    \hat{v}(s, \mathbf{w}_{*}) = \min_a \sum_{s'} \Pr(s'|s,a) \left[ H(s') + \hat{v}(s', \mathbf{w}_{*}) \right] \, .
\end{equation}
 Solving for the optimal weights $\mathbf{w_{*}}$ involves minimizing the residual error, known as the Bellman optimality error~\eqref{eq:bellman_error}~\cite{loisy_searching_2022}
\begin{equation} \label{eq:bellman_error}
\begin{split}
    L(\mathbf{w}) = \mathbb{E}_s & \Biggl[ \min_a \sum_{s'} \Pr(s' | s, a) 
     \\ 
    & \left[ H(s') + \hat{v}(s', \mathbf{w}) \right] - \hat{v}(s, \mathbf{w}) \Biggr]^2 \, .
\end{split}
\end{equation}
The expectation is taken over belief states~$s$ visited when following policy~$\hat{\pi}$ derived from $\hat{v}$ and defined by~Eq.~\eqref{eq:rlpolicy}
\begin{equation}
    \hat{\pi}(s, \mathbf{w}) = \arg \min_a \sum_{s'} \Pr(s'|s,a) \left[ H(s') + \hat{v}(s', \mathbf{w}) \right] \, . \label{eq:rlpolicy}
\end{equation}
The functional~$L(\mathbf{w})$ is the \emph{loss function}, and \emph{training} the neural network entails optimizing the weights $\mathbf{w}_\star=\arg\min_\mathbf{w} L(\mathbf{w})$ using stochastic gradient descent, aiming to minimize the loss function. The subsequent DRL policy is given by Eq.~\eqref{eq:rlpolicy} evaluated with the optimized weights $\pi_{\textrm{DRL}}(s, \mathbf{w}_{*})=\hat{\pi}(s, \mathbf{w}_\star)$. Note that by initializing the weights of the value function in proximity to zero, the untrained DRL policy~\eqref{eq:rlpolicy} closely resembles the infotaxis policy~\eqref{eq:infotaxis_policy_2}.

Our DRL algorithm is based on Deep Q-Networks (DQN)~\cite{mnih_human-level_2015}, an extension of Q-learning to deep neural networks. While most RL algorithms are model-free and use state-action values $q(s,a)$, our study is inspired by the approach of \citeauthor{loisy_searching_2022}~\cite{loisy_searching_2022} and \citeauthor{loisy_deep_2023}~\cite{loisy_deep_2023} (with accompanying code~\cite{otto_code}), and adopts a model-based approach due to our exact knowledge of transition probabilities $\Pr(s',r|s,a)$. This allows us to work with $\hat{v}(s, \mathbf{w})$ instead of $\hat{q}(s,a,\mathbf{w})$.

\begin{figure*}[tb]
    \centering
    \includegraphics[width=\textwidth]
    {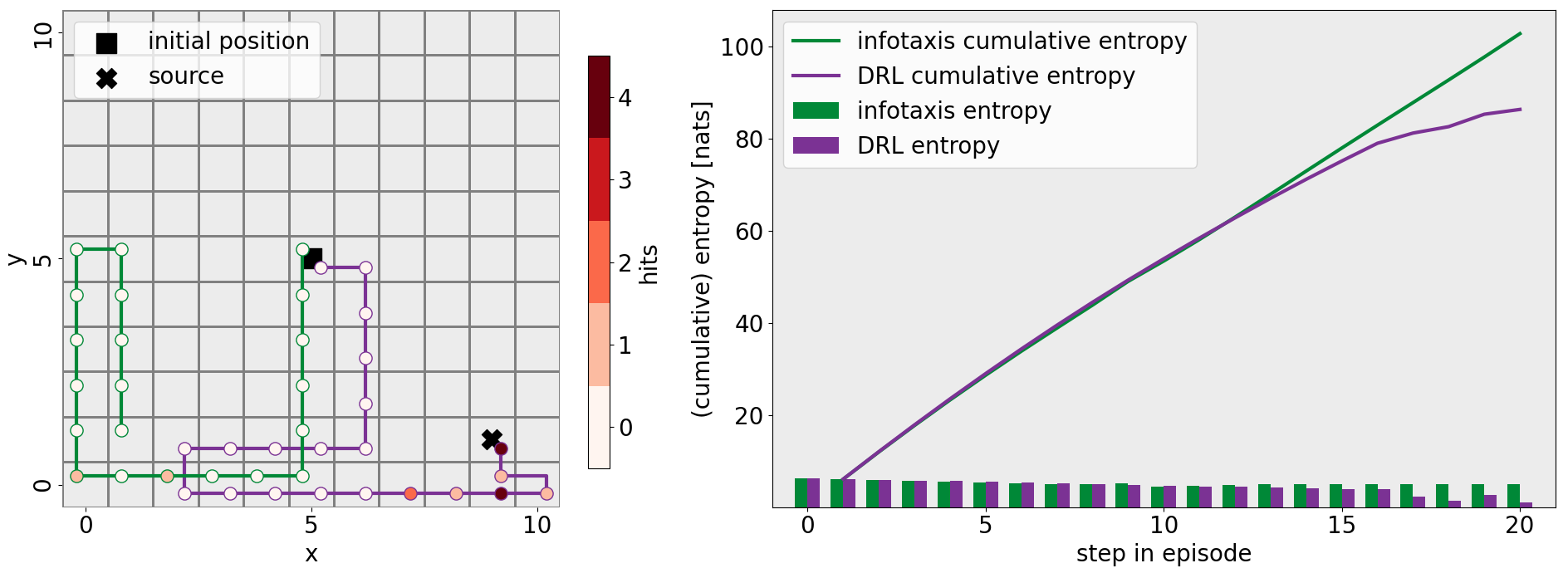}
    \caption{Agent navigation in an environment with $\tilde{V} = 2$ and $\tilde{D} = 2$, and source term parameters $\tilde{\mathbf{x}}_{\mathrm{s}} = (9,1)$ and $\tilde{\phi} = 2$: (left) DRL and infotaxis sampling paths, and (right) (cumulative) entropy. Lower entropy indicates a reduced uncertainty in the estimation of the source term parameters.}
    \label{fig:paths}
\end{figure*}

We utilize a feedforward neural network and compare two different network architectures: (a)~an architecture with 3 fully connected hidden layers each containing 32 units, and (b)~an architecture with 4 hidden convolutional layers with a filter size of 32 and a kernel size of 3. Average pooling with kernel size 2 and stride 2 is applied. Both network architectures use rectified linear units (ReLU) for activations, and a linear output layer. The network sizes are tuned such that adding more weights does not increase their performance any further. The weight parameters are trained over 20,000 episodes using stochastic gradient descent with mini-batches of size 128. 

During preprocessing, we transform belief state~$s = (\mathbf{x}, \Pr(\boldsymbol{\theta}))$ containing the agent's position and source term probability distribution ($N_{x} \times N_{y} \times N_{\phi}$ tensor), into a source term probability distribution centered on the agent ($(2 N_{x} -1) \times (2 N_{y} -1) \times N_{\phi}$ tensor)~\cite{otto_code}. Tensor elements outside the physical domain are set to zero. When used as an input for architecture (b), the tensor can be conceived as an image representing spatial data with dimensions $(2 N_{x} -1) \times (2 N_{y} -1)$ pixels and $N_{\phi}$ channels. The tensor is flattened when used as an input for architecture~(a).

\section{Results}~\label{sec:results}

We adopt a non-dimensional representation of observation model~\eqref{eq:plume_model} whereby hit counts are governed by dimensionless parameters: source flux $\tilde{\phi}$, location $\tilde{\mathbf{x}}$, agent radius $\tilde{r}$, gas lifetime $\tilde{\tau}$, mean wind speed $\tilde{V}$, and effective diffusivity $\tilde{D}$. We assume that the source is located within a domain of 11 x 11 grid cells and has one of the following fluxes: $\tilde{\phi} = {\{1, 2, 3, 4, 5\}}$. As a result, we consider 605 possible source term scenarios. The constants $\tilde{r} = 0.5$ and $\tilde{\tau} = 10,000,000$ are fixed. In sections~\ref{sec:results_NN} and~\ref{sec:results_info_vs_RL}, we maintain $\tilde{V} = 2$ and $\tilde{D} = 2$, while other values for $\tilde{V}$ and $\tilde{D}$ are explored in section~\ref{sec:results_wind}. The agent starts in the center of the domain and performs 20 actions. The evaluation statistics presented herein are derived from 5,000 episodes. 

\subsection{Deep neural network}~\label{sec:results_NN}

We assess the performance of a DRL policy approximated through a fully connected~(FC) architecture and a convolutional neural network~(CNN) as outlined in section~\ref{sec:DRL}. The FC architecture attains an average cumulative entropy of $66~\textrm{nats}$, while the CNN performs slightly better with an average of $65~\textrm{nats}$. In this study, we define \emph{success} as the agent's ability to estimate the correct source location and flux with a final posterior probability of $\Pr \geq 0.5$. The FC architecture achieves success in $70\%$ of episodes, while the CNN achieves this in $71\%$ of episodes. The CNN architecture is used in subsequent sections of this study. 

\begin{figure}[tb]
    \centering
    \includegraphics[width=\linewidth]{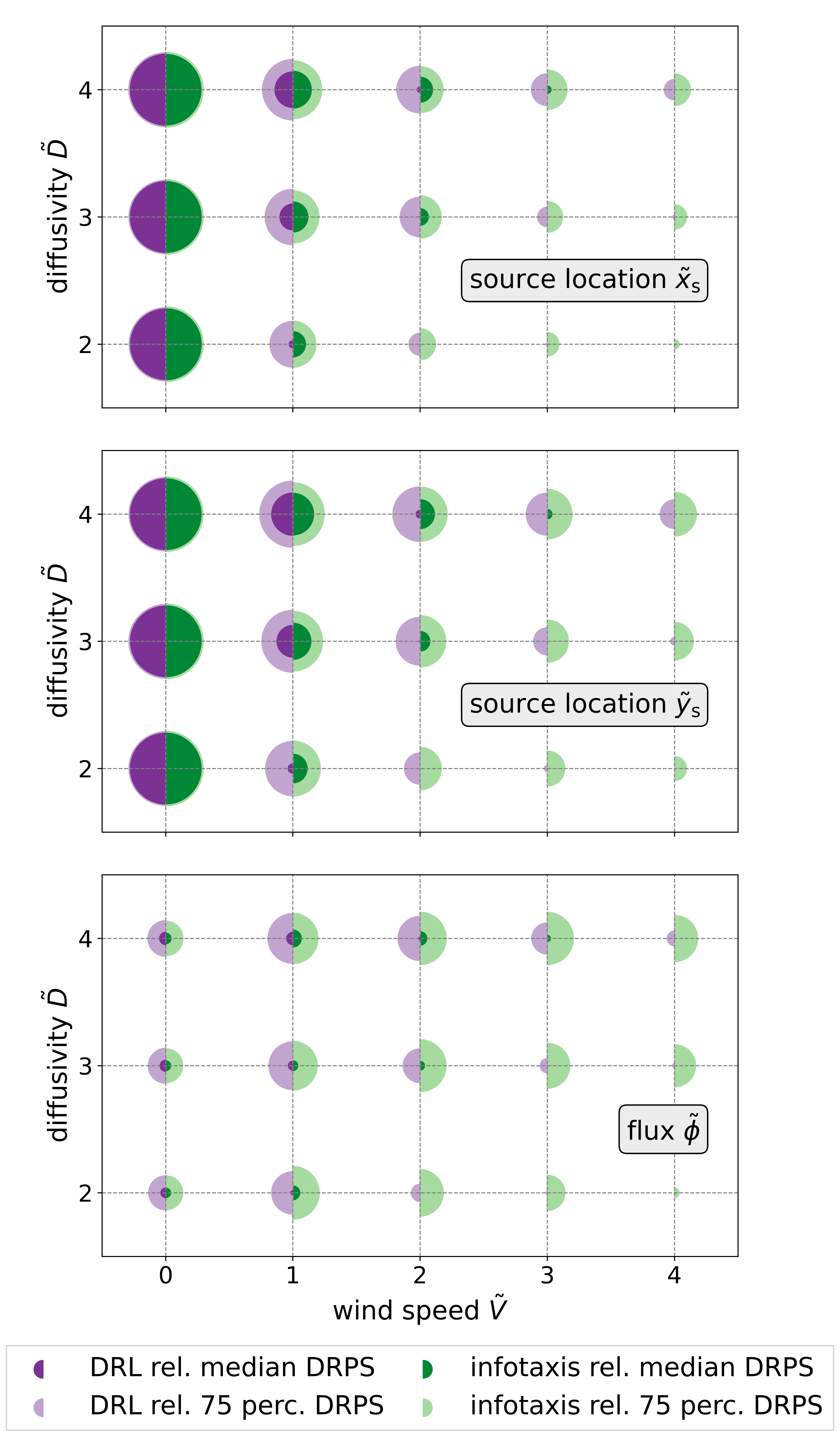}
    \caption{Relative Discrete Ranked Probability Score~(DRPS) for source location $\tilde{x}_{s}$ (top), $\tilde{y}_{s}$ (middle) and flux $\tilde{\phi}$ (bottom). Circle radii correspond to the median (dark shade) and 75$^{\textrm{th}}$ percentile (light shade) relative DRPS. DRL results are shown as (purple) left half circles and infotaxis results are shown as (green) right half circles.}
    \label{fig:wind_speed}
\end{figure}

\subsection{DRL versus infotaxis}~\label{sec:results_info_vs_RL}

We observe that the infotaxis policy achieves success in $58\%$ of episodes (compared to 71\% for DRL). Figure~\ref{fig:paths} shows a scenario with two agents navigating an environment with a gas source with $\tilde{\phi} = 2$ at $\tilde{\mathbf{x}}_{\mathrm{s}} = (9,1)$: one agents follows the infotaxis policy, while the other adheres to the DRL policy. In the absence of receiving any hits, both agents move downwind (negative $y$-direction). The DRL agent follows a sweeping pattern downwind, whereas the infotaxis agent traverses along the edge of the domain. Upon detecting hits, the entropy of the agent's belief state reduces more significantly compared to zero hit events. Over the entire episode, the DRL agent receives more valuable information than the infotaxis agent, leading to a greater reduction in uncertainty of its belief state, and subsequently a superior (i.e. smaller) cumulative entropy. 

We investigate whether DRL outperforms infotaxis across all source term scenarios considered in this study. As demonstrated in Table~\ref{tab:per_flux}, the success rate increases with source strength, consistently favoring the DRL approach over infotaxis. By evaluating the performance across different source locations (not shown), we find that the DRL policy surpasses the success rate of infotaxis in $76\%$ of source locations, primarily in the upper and right segments of the domain. Conversely, infotaxis outperforms DRL in $24\%$ of source locations, mainly in the lower-left segment of the domain. As demonstrated in Figure~\ref{fig:paths}, this coincides with the region that the infotaxis agent navigates towards in the absence of any detected hits.

\begin{table}[tb]
    \centering
    \caption{Success rate of DRL and infotaxis evaluated across different flux values~$\tilde{\phi}$ (training encompassed all source term scenarios).} 
    \label{tab:per_flux} 
    \renewcommand{\arraystretch}{1.5}
    \begin{tabular}{@{}l|c c c c c} 
         {} & $\tilde{\phi}=1$ & $\tilde{\phi}=2$ & $\tilde{\phi}=3$ & $\tilde{\phi}=4$ & $\tilde{\phi}=5$ \\
         \hline
         \hline
         DRL   &        {$40\%$}  & {$65\%$} & {$80\%$} & {$84\%$} & {$85\%$} \\
         infotaxis  &   {$31\%$}  & {$49\%$} & {$65\%$} & {$72\%$} & {$76\%$} \\
    \end{tabular}
\end{table}

\subsection{Sensitivity to $\tilde{V}$ and $\tilde{D}$}~\label{sec:results_wind}

Figure~\ref{fig:wind_speed} illustrates a relative Discrete Ranked Probability Score~\cite{candille_evaluation_2005} (DRPS$_{\textrm{posterior}}$ / DRPS$_{\textrm{prior}}$) for DRL and infotaxis policies across environments with varying known diffusivities~$\tilde{D}$ and wind speeds~$\tilde{V}$. The agents are trained under these specific environmental conditions. DRPS is a measure for belief state accuracy, favoring lower scores. Note that DRPS$_{\textrm{prior}}$ is the initial uniform belief state and remains consistent across all test cases. We observe improved estimations of source location in conditions with higher winds. In cases where $\tilde{V}>0$, DRL outperforms infotaxis. Additionally, we observe that the estimation of source location on the crosswind axis~($\tilde{x}_{\mathrm{s}}$) surpasses that on the along windward axis~($\tilde{y}_{\mathrm{s}}$), and that lower diffusivities are beneficial for STE in such cases. Under isotropic conditions ($\tilde{V}=0$), DRL and infotaxis perform comparably given the specified training parameters. 

\section{Discussion}~\label{sec:discussion}

The findings presented in this study are a first effort in applying DRL with an information-based reward to solve the STE problem with mobile gas sensors. While infotaxis adopts a myoptic perspective, maximizing information gain solely for the upcoming action, DRL has the capability to develop a far-sighted strategy for action selection, considering information gain across the entire sampling trajectory. This study demonstrates that adopting a DRL approach has the potential to improve the accuracy of STE. 

Our findings show that DRL outperforms infotaxis in non-isotropic plume environments. The policy allows the agent to select actions that might not be immediately optimal, but are chosen in anticipation of achieving better rewards in the future. Comparative analysis shows that the DRL agent opts to stay (downwind) centralized when its belief over the source term parameters remains uncertain. In contrast, the infotaxis agent tends to move towards the edge of the domain under similar circumstances. This trajectory could bring the infotaxis agent closer to the source and valuable information, but also poses a risk of leading the agent in the opposite direction, further away from the source and information about the source term parameters. 

DRL's superiority over infotaxis does not extend to isotropic plumes in this study, possibly due to the increased symmetry of the hit map and hence limited directional information. In contrast, recent work by \citeauthor{loisy_searching_2022}~\cite{loisy_searching_2022} showed DRL's efficiency in source localisation within isotropic plumes. Key distinctions between our studies are that \cite{loisy_searching_2022} exclusively focuses on localisation, omitting flux estimation. In their case, the flux strength was known which assisted the agent in source seeking, whereas in our study the search for informative observations is further complicated by the unknown source strength.
Their policy employed a $-1$ reward, terminating episodes upon the agent's arrival at the source - information not available to our agent. Moreover, episodes initiated with the agent receiving a hit, while episodes can start with zero hit observations in our study.

\citeauthor{loisy_searching_2022}~\cite{loisy_searching_2022} find that the performance of space-aware infotaxis is close to the optimal policy obtained by DRL for source seeking. This variation on traditional infotaxis incorporates an expected distance measure between agent and source, and is designed to efficiently select the shortest path to the source. Our approach differs: we aim to strategically select a 20-step path that optimally constrains the source term parameters through observations. In the environment studied currently, the optimal policy indeed leads the agent towards the source. Nonetheless, unlike \citeauthor{loisy_searching_2022}~\cite{loisy_searching_2022}, our agent entertains multiple possible flux strengths, such that the agent may place equal probability to the source being close by with a low flux strength as it being further away with a higher flux strength. 
Furthermore, we deliberately refrain in this study from rewarding actions that select the shortest path upfront, as our long-term aim is to develop a framework that can perform in various environments. In particular, future work will include domains with multiple sources. In this case, we anticipate that the shortest path to one of the sources is not the optimal policy to constrain the uncertainty over the parameters characterising all sources.

\section{Outlook}

As we look ahead to the practical implementation of our DRL framework, we identify several challenges and possibilities: 
(a)~The framework presented in this study is trained for known wind conditions and diffusion. Real-world application requires training across a range of atmospheric conditions, which can be inferred from observations by mobile and ground-based sensors, or a combination of the two~\cite{pirk_2022}. 
(b)~The current observation model~\eqref{eq:plume_model} assumes steady-state atmospheric conditions, a simplification that may not hold in real-world scenarios. More complex, time-dependent models, as employed by \citeauthor{pirk_2022}~\cite{pirk_2022}, may enhance the accuracy of the inference results. However, these models are computationally expensive, posing challenges for online usage. 
(c)~We currently assume the presence of a single source within the domain. However, in reality, each ``grid cell'' represents a possible greenhouse gas source or sink. Future research aims to extend the framework to map heterogeneous greenhouse gas fluxes across a domain. 
(d)~Applying the framework in field studies requires upscaling to larger domains and parameter spaces. Future work should investigate whether the FC and CNN architectures in this study can sustain their comparable performance within a realistic computational budget. 

Despite the anticipated challenges in implementing the framework in real-world scenarios, our study represents a significant advancement in the methodology for STE of gas sources. It demonstrates encouraging outcomes achieved through the guidance of drones by information gain strategies. 

\section*{Acknowledgements}
This work was supported by the Research Council of Norway (project 301552 ``Upscaling hotspots – understanding the variability of critical land-atmosphere fluxes to strengthen climate models (Spot-On)'' and project 333232 ``Strategies for Circular Agriculture to reduce GHG emissions within and between farming systems across an agro-ecological gradient (CircAgric-GHG)''). This work is a contribution to the strategic research initiative LATICE (Faculty of Mathematics and Natural Sciences, University of Oslo, project UiO/GEO103920) as well as the Centre for Computational and Data Science (dScience, University of Oslo).

\printbibliography

\end{document}